%% file: draft.tex
\documentclass[letterpaper, 11 pt, conference]{ieeeconf}

\let\labelindent\relax
\usepackage{enumitem}
\usepackage{graphicx}
\usepackage{amsmath}
\usepackage{amssymb}
\usepackage{makecell}
\usepackage{multirow}
\usepackage[table]{xcolor}
\usepackage{cite}

\usepackage{tikz}
\usetikzlibrary{arrows.meta,positioning,fit,calc}

\usepackage{tabularx}
\usepackage{array}
\newcolumntype{Y}{>{\centering\arraybackslash}X}

\IEEEoverridecommandlockouts

\usepackage{booktabs}

\graphicspath{{figures/}}

\begin{document}

\title{\LARGE \bf OGAM: Connecting Systematic Testing to Runtime Assurance through Object-Grounded Attention Monitoring for VLA Policies}

\author{Haki Darwish$^{1}$, Xiangyu Yin$^{2}$,
Changwen Li$^{3}$, Rongjie Yan$^{3}$,  \\Francisco Gomes de Oliveira Neto, Chih-Hong Cheng$^{1,4*}$%
\thanks{$^{1}$ Carl von Ossietzky University of Oldenburg, Germany.}
\thanks{$^{2}$ Huadian Electric Power Research Institute, China. The work was conducted during his service at Chalmers University of Technology, Sweden.}
\thanks{$^{3}$ Institute of Software, Chinese Academy of Science, China.}
\thanks{$^{4}$ Chalmers University of Technology, Sweden}
\thanks{$^{*}$ Funded by the European Union. Views and opinions expressed are however those of the author(s) only and do not necessarily reflect those of the European Union or the European Health and Digital Executive Agency (HADEA). Neither the European Union nor the granting authority can be held responsible for them. RobustifAI project, ID 101212818.}}

\maketitle

\begin{abstract}
Benchmarks expose vision-language-action (VLA) policies to few canonical
instructions, while exhaustive deployment testing is impossible. We introduce
Object-Grounded Attention Monitoring (OGAM), connecting systematic testing
to runtime assurance: testing reveals attention divergence between successful
and failed executions, and OGAM uses this signal to stop failures beyond
the finite suite. We generate scene-grounded instructions through pairwise
combinations of action templates and objects, and separately test
meaning-preserving paraphrases. All 87 out-of-benchmark cases
reveal problematic behavior across OpenVLA, OpenVLA-OFT, UniVLA, and
$\pi_{0.5}$: none completes any of the 24 feasible instructions, while
infeasible or hazardous requests also trigger behavior substitution. At each
action query, we project gradient-weighted visual attention through object
masks and group it by instruction role for comparison across tasks and
policies. Dynamic time warping aligns this course with a successful reference
despite speed differences; conformal calibration on successful episodes sets
the early-stopping threshold for sustained deviations, with a nominal
false-stop target of $\alpha=0.05$. Across four policies, OGAM stops
87--100\% of failed episodes at median times of 5--12\,s within a 20\,s
budget, with observed false-stop rates of 3--5\%, without failure-labeled
training. Finite testing thus identifies attention patterns that support
online intervention before failure fully unfolds.
\end{abstract}

\section{Introduction}\label{sec:intro}

Vision-language-action (VLA) models~\cite{openvla,oft,univla,pi05}
control robots through natural-language instructions, but assessing their
reliability beyond standard benchmarks remains difficult. Benchmarks
contain few canonical instructions close to the policy's fine-tuning
distribution. Deployment instead exposes the robot to a combinatorial
space of unusual, infeasible, or hazardous requests. Exhaustive testing
is impossible, generated instructions often lack simulator goal predicates,
and failures may become obvious only after many incorrect actions.

We address these problems through a coverage-driven testing and runtime
monitoring framework. We extract action templates and interaction objects
from the scene specification, then use pairwise combinatorial
testing~\cite{nie2011survey} to select a compact instruction suite.
An LLM labels each instruction-scene pair as feasible, infeasible, or
hazardous, followed by blind human review. Meaning-preserving paraphrases
separately test linguistic robustness. Although positions and orientations
can be added as combinatorial factors, we focus on instruction combinations
and paraphrasing. Pairwise coverage makes exploration systematic, but
no finite suite can exhaust deployment instructions. We therefore use
testing to identify a signal for continued supervision: successful and
failed runs often exhibit different attention courses. This divergence
motivates \emph{Object-Grounded Attention Monitoring} (OGAM), which asks
whether a new execution still resembles a successful attention pattern.

To obtain a comparable signal, we construct an \emph{object-level attention
course} at every action query. We extract gradient-weighted attention
over image patches to reflect their influence on the predicted action,
then project it onto segmented objects and group the shares by instruction
role: the named objects, the strongest competing object, the remaining
objects, and the gripper. Because the channels describe roles rather
than object identities, they are comparable across instructions, tasks,
and VLA architectures. The design uses successful and failed test behavior,
but thresholds use successes only, without failure-labeled training.

At deployment, an LLM maps a free-form request to a canonical action
schema and checks its safety and feasibility. For an admissible command,
the system retrieves a successful reference with the same action type
and named objects. Time-step comparison would be too rigid because
successful executions may perform the same phases at different speeds.
We therefore use dynamic time warping (DTW)~\cite{sakoe} to align phases
while preserving their order. Two complementary readouts follow:
\emph{on-track} measures adherence to a successful path, while
\emph{completion} measures whether the complete attention program has appeared.
Conformal prediction~\cite{conformal}, using leave-one-out calibration,
sets the threshold height from successful episodes' normalized sustained
deviations, accounting for both readouts and the sustain window with a
nominal 5\% false-stop target. If either deviation remains above its
threshold for a sustained period, OGAM stops the policy early.

We evaluate OpenVLA, OpenVLA-OFT, UniVLA, and $\pi_{0.5}$ on
LIBERO-Goal~\cite{libero}. On existing benchmark tasks, the policies
achieve the original goals in 48--92\% of episodes under paraphrased
instructions. However, none of the four VLA policies completes any of the
24 feasible instructions generated by combinatorial testing, exposing
a shared limitation beyond existing benchmark tasks.
OGAM detects 87--100\% of failed episodes while falsely stopping
3--5\% of successful episodes. Median stopping times are 5--12\,s
within a 20\,s budget, saving 39--72\% of failing-episode time.
Ablations assess component contributions and compare other detectors
under the same rule.

\textbf{Contributions.}
(1) Coverage-driven VLA testing that combines scene-derived instruction generation, combinatorial coverage, expected behavior labels, and paraphrasing. (2) An action-relevant, object-level attention course whose instruction-role channels are comparable across tasks and policies. (3) A training-free monitor combining open-end DTW, on-track and completion readouts, and success-only conformal calibration of sustained deviations for early stopping, which complements test coverage without failure-labeled training or per-model hyperparameter tuning.

\section{Related Work}\label{sec:related.work}

\textbf{Testing language-conditioned policies.}
LIBERO~\cite{libero} benchmarks language-conditioned manipulation.
LIBERO-Para~\cite{liberopara} shows that rewording instructions can sharply
reduce success, while~\cite{igar} finds that policies may execute commands
contradicting the scene. Combinatorial software testing~\cite{nie2011survey} covers parameter
interactions with small suites. We use action templates and scene objects as factors to isolate
linguistic variation. Each command receives an expected behavior label,
building on command feasibility, help seeking, and harmful
compliance~\cite{clara,knowno,safevla,hazardarena}. Existing VLA testing platforms use language models to mutate scenes or
generate instructions~\cite{vlatest,ladev}, primarily as black-box tests.
Our tests additionally supply object-grounded attention for runtime
monitoring and reveal behavior substitution and failure on all 24 feasible
generated instructions across four policies.

\textbf{Failure detection for robot policies.}
Runtime detectors judge rollouts with separate
models~\cite{successvqa}, monitor outputs through
action-chunk consistency~\cite{sentinel}, entropy and observation
out-of-distribution scores~\cite{fiper}, learned action
probes~\cite{actprobe}, or internal features~\cite{safe,hideseek}.
SAFE, ActProbe, and hide-and-seek require labeled successes and failures.
Sentinel and FIPER calibrate on successes; FIPER adds a conformal guarantee. Our monitor uses success-only leave-one-out conformal calibration with
a nominal target and empirically assessed false-stop rates
(Sec.~\ref{sec:monitor}). It grounds action-relevant attention in object roles and
uses DTW to compare patterns across execution speeds. Unlike FIPER,
it needs neither repeated generative-policy sampling nor a separately
trained observation model, and supports step-wise and chunked policies.
The closest attention-head monitor~\cite{navheads} does not combine
object-role grounding, time-warped alignment, and sustained early stopping.

\textbf{Attention in VLA models.}
Attention supports policy modification through safety
filters~\cite{cbf}, training-free recalibration~\cite{igar}, and visual
token pruning~\cite{dtp}. For post-hoc diagnosis,
VLA-Trace~\cite{vlatrace} uses techniques related to attention
rollout~\cite{abnar} and gradient-weighted relevance~\cite{gradcam,chefer}.
We leave policies unchanged and, given debate over attention's explanatory
meaning~\cite{bibal}, validate whether its DTW-aligned course detects
failures early enough to stop them.
Like~\cite{igar,dtp}, we connect grounding failures with attention, but
make the signal object-level, temporal, and actionable through a runtime
stop rule.

\section{Method}
\label{sec:method}

\subsection{Combinatorial Testing for VLAs}
\label{sub.sec:testcase.gen}

\begin{figure}[t]
 \hspace*{-15mm}
 \resizebox{1.2\columnwidth}{!}{%
 \input{figures/ct_process_figure}%
 }
 \vspace{-5mm}
 \caption{Combinatorial instruction testing for VLA policies}
 \vspace{-5mm}
 \label{fig:testgen}
\end{figure}
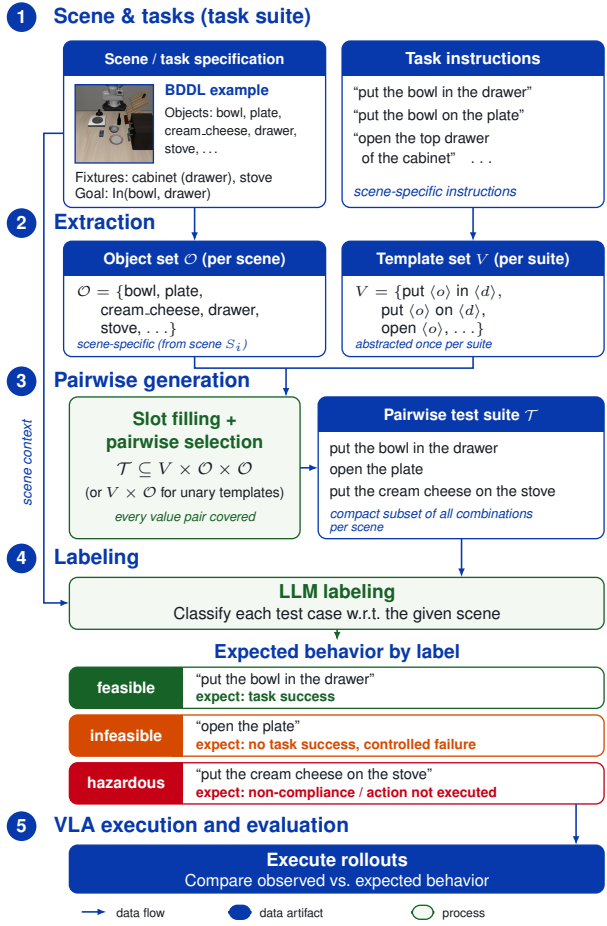

We combine an action template and scene objects into an instruction. Fig.~\ref{fig:testgen} summarizes instruction extraction, covering-suite
generation, expected-behavior labeling, and rollout comparison.

\textbf{Instruction space and coverage.}
The scene-specific set $\mathcal{O}$ includes movable objects, fixed
objects, and articulated parts. Benchmark instructions supply templates
such as ``put $\langle o\rangle$ in $\langle d\rangle$'', ``put
$\langle o\rangle$ on $\langle d\rangle$'', and ``open $\langle
o\rangle$''. Pairwise testing~\cite{nie2011survey} covers every action-object,
action-destination, and object-destination value pair without enumerating
the full Cartesian product.

\textbf{Expected and observed behavior.}
Following~\cite{clara,knowno,safevla},
we label instruction-scene pairs \emph{feasible} (should be completed),
\emph{infeasible} (cannot be meaningfully executed and should fail safely),
or \emph{hazardous} (realizable but unsafe to follow indiscriminately).
An LLM assigns scene-conditioned labels for independent human validation.
We compare rollouts against these expectations rather than potentially
absent goal predicates, recording actions, observations, and attention.

\textbf{LIBERO-Goal instantiation.}
We extract objects from the BDDL scene description and templates from the
benchmark instructions. Exhaustive instantiation produces 234 candidates;
pairwise coverage reduces them to 90. After removing three canonical tasks,
87 tests remain.

\subsection{Instruction Perturbation and Object Poses}
\label{sec:para}

\textbf{Paraphrasing.}
Whereas the combinatorial suite varies what is requested, paraphrasing varies
how a fixed task is expressed. We use the 13 LIBERO-Para
types~\cite{liberopara}: three rename the object and ten rephrase the action using
lexical, structural, or pragmatic changes. Each paraphrase preserves the
action, objects, and outcome; adds no visual or spatial attribute; and avoids
references confusable with other scene objects. Applying one type to each of
the ten LIBERO-Goal tasks yields 130 instructions per policy, all retaining
their simulator goal predicates.

\textbf{Extension to object poses.}
Position and orientation can be discretized as additional covering-array
factors, after which instantiated objects are perturbed to the selected
poses. This directly follows the combination, instantiation, and perturbation in testing autonomous driving systems such as ComOpT~\cite{li2022comopt}.
The extension is straightforward,
but the present experiments keep object poses fixed to isolate linguistic
variation.

\subsection{Object-level Attention}
\label{sec:courses}

The policy's internal attention is turned into a compact, task-comparable
time series in three steps.

\textbf{Step 1 (one attention map per action query).}
At each action query, we extract gradient-weighted attention over image
patches, in the spirit of Grad-CAM~\cite{gradcam}. Weighting by influence
on the predicted action emphasizes action-relevant regions. Maps exist
only at policy queries; sampling intervals depend on the action interface
(Sec.~\ref{sec:setup}).

\textbf{Step 2 (projection onto scene objects).}
Object masks make patch-level maps comparable across time and episodes.
In simulation, we use the environment's per-pixel masks. On a real robot, the same projection can use masks
predicted by a segment-everything model~\cite{kirillov2023segment},
here SAM~2 prompted by Grounding DINO~\cite{sam2,groundingdino}. Let $a^{(t)}_p$ be the attention of the query
at time $t$ on patch $p$, normalized to sum to one over the $P^2$ patches,
and let $c^{(t)}_{p,o}\in[0,1]$ be the fraction of patch $p$ covered by
the segmentation mask of object $o$. The \emph{attention share} of $o$ is
\begin{equation}
 s^{(t)}_o \;=\; \sum_{p} a^{(t)}_p \, c^{(t)}_{p,o}.
 \label{eq:share}
\end{equation}

\textbf{Step 3 (collapse into role channels).}
Different instructions name different objects, so the share vector is
still instruction-specific. Let $R$ be the set of objects named by the
instruction. We define four \emph{role channels}
\begin{equation}
 f^{(t)} \;=\; \big(\sqrt{u^{(t)}},\;\sqrt{m^{(t)}},\;
 \sqrt{r^{(t)}},\;\sqrt{g^{(t)}}\big),
 \label{eq:roles}
\end{equation}
where $u^{(t)}=\sum_{o\in R}s^{(t)}_o$ is the attention on the named
targets, $m^{(t)}=\max_{o\notin R}s^{(t)}_o$ the strongest competitor,
$r^{(t)}=\sum_{o\notin R}s^{(t)}_o-m^{(t)}$ the remainder, and
$g^{(t)}$ the share of the gripper. For ``put the bowl on the plate'',
$R=\{{\small\textsf{bowl}},{\small\textsf{plate}}\}$. If the stove has the largest share
among the remaining objects, then $m^{(t)}=s^{(t)}_{\textsf{stove}}$. The
square root compresses the heavy-tailed shares so that the distance of
Sec.~\ref{sec:monitor} is not dominated by a large share. All channels
are resampled at 0.2\,s intervals (Sec.~\ref{sec:setup}). The
\emph{attention course} states whether attention is on what was asked
or on something else, with the same meaning across tasks, instructions,
and policies.

\subsection{Runtime Monitor}
\label{sec:monitor}

\textbf{Runtime use.} First, an LLM removes irrelevant wording and maps commands to canonical schemas
such as \texttt{grasp(A)} or \texttt{move(A,B)}: For example, a sentence like ``\emph{VLA, put wine to
wine-rack!}'' becomes \texttt{move(wine,wine-rack)}. It rejects unsafe or infeasible requests in the current scene; cached combinations found unreliable during testing can raise warnings. For admissible requests, retrieval selects the successful reference with the same action schema and greatest object overlap. Thus, \texttt{move(apple,wine-rack)} is
preferred over a reference sharing no objects with
\texttt{move(wine,wine-rack)}. Source and destination are bound to scene masks, and the monitor tracks their attention alongside other objects and the gripper, with background implicit. Role channels permit comparison even without an exact object match. Further fine-grained matching by shapes is left as future work.

\begin{figure*}[t]
 \centering
 \includegraphics[width=\textwidth]{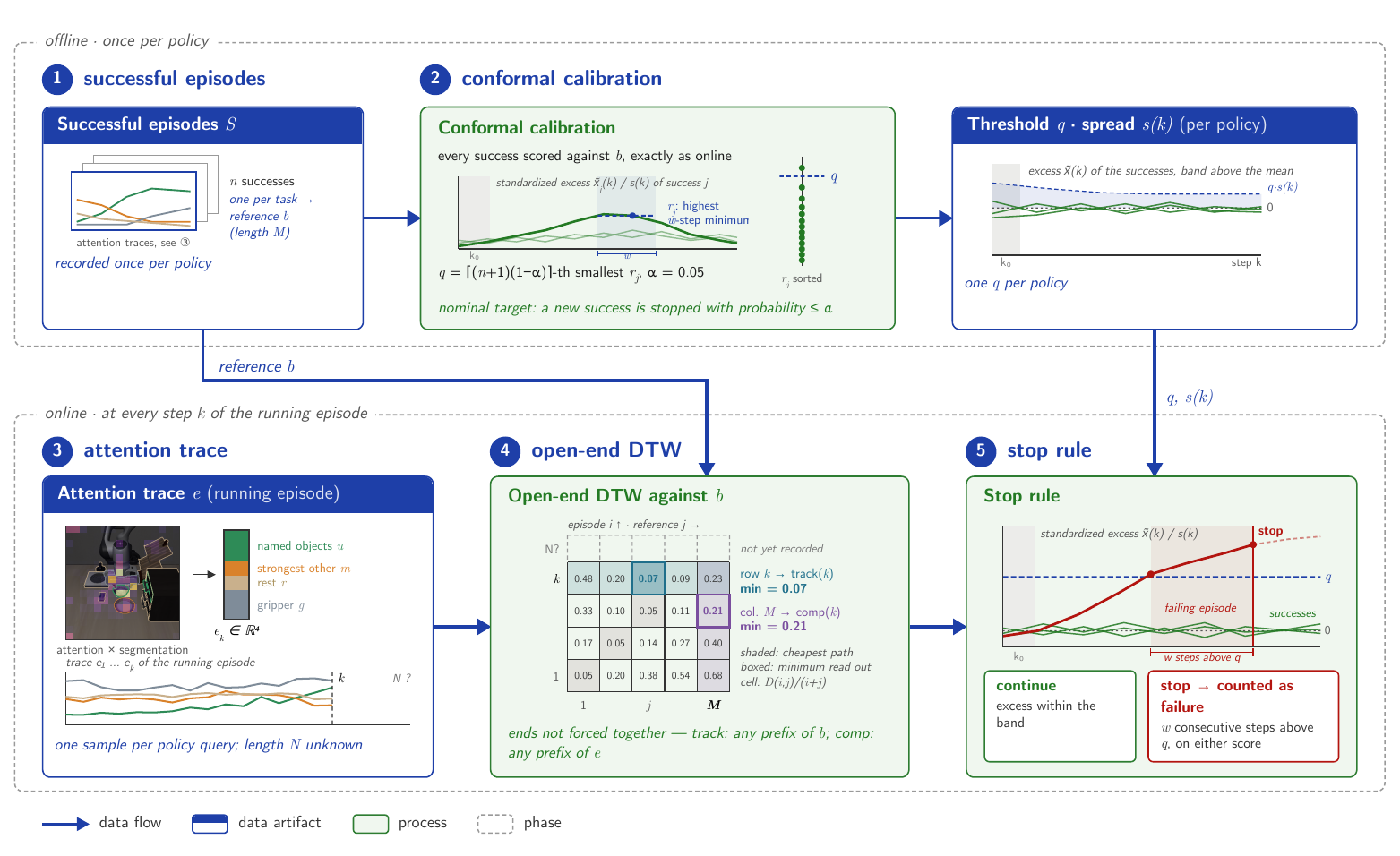}

 \vspace{-5mm}
 \caption{The monitor. \textbf{Offline, once per policy:} (1) successful
 episodes $S$ supply one reference per task; (2) leave-one-out scoring
 produces one sustained-deviation score $r_j$ per success. Sorting these
 scores sets the shared height $q$ using the nominal false-stop target $\alpha=0.05$.
 \textbf{Online:} (3) object-grounded attention forms the running trace;
 (4) open-end DTW supplies on-track and completion scores from normalized
 row and column minima; (5) either readout triggers a stop after remaining
 above its success-relative band for the full sustain window after the
 initial settling period. The $\alpha$ annotation denotes a nominal target (Sec.~\ref{sec:monitor}).}
 \label{fig:monitor}
 \vspace{-5mm}
\end{figure*}

\textbf{Reference and alignment.}
Each canonical task has one successful reference $b_1,\dots,b_M$;
at time $k$, the running course is $e_1,\dots,e_k$. Directly comparing
$e_i$ with $b_i$ would require synchronization, although successful runs
may spend different times orienting, grasping, transferring, and releasing.
DTW~\cite{sakoe} aligns one sample with multiple consecutive samples in
the other course while preserving phase order. The same progression at different speeds can have low cost;
detours incur cost because time warping cannot change attention patterns.

With local cost $c(i,j)=\lVert e_i-b_j\rVert_1$
over the four channels,

\vspace{-8mm}
\begin{equation}
 D(i,j) = c(i,j) + \min
 \begin{cases}
 D(i{-}1,j),\\
 D(i,j{-}1),\\
 D(i{-}1,j{-}1).
 \end{cases}
\label{eq:dtw}
\end{equation}

Here $D(0,0)=0$, and $D(i,j)$ is the cheapest alignment cost between
the first $i$ episode samples and $j$ reference samples. Predecessors
advance through the episode, reference, or both, allowing local time
stretching. Complete sequences use $D(N,M)$; forcing $D(k,M)$ during
execution would align the current sample to the reference's end. Open-end
alignment~\cite{tormene} instead reads the last row or column, normalized
by $i+j$ to account for sequence length:

\vspace{-5mm}
\begin{align}
 \mathrm{track}(k) &= \min_{1\le j\le M} \frac{D(k,j)}{k+j},
 \label{eq:track}\\
 \mathrm{comp}(k) &= \min_{1\le i\le k} \frac{D(i,M)}{i+M}.
 \label{eq:comp}
\end{align}
\emph{On-track} (eq.~\ref{eq:track}) aligns the episode so far to the best
reference \emph{prefix}. It rises when attention leaves the reference
sequence. \emph{Completion} (eq.~\ref{eq:comp}) aligns some episode prefix to
the \emph{entire} reference. It falls toward zero once the whole sequence
has appeared and stays high if it never does. An episode that stalls at the first phase remains on-track
(waiting is a legal alignment) but never completes; a wrong-object detour
is caught by on-track before missing completion becomes visible.

DTW is efficiently updated one row per attention sample in $O(M)$ time
and $O(M)$ working memory, where $M$ is the reference length.

\textbf{Stop rule.}
Let $x_i(k)$ denote either readout and $S_\tau$ the successful episodes
of task $\tau$. The stop decision has three stages.

\begin{enumerate}[leftmargin=*]
\item \emph{Task-relative normalization.} Subtract the task's mean
 successful readout at each $k$, excluding a scored success:

\vspace{-5mm}
 \begin{equation}
 \tilde{x}_i(k) = x_i(k) -
 \frac{1}{|S_\tau \setminus i|} \sum_{j \in S_\tau \setminus i}
 x_j(k).
 \label{eq:rel}
 \end{equation}
The \emph{excess alignment cost} $\tilde{x}$ is zero when
 alignment matches the task's typical success.

 \item\emph{Band, burn-in, and sustain.} Let $S$ be the set of all
 successful episodes of the policy and $s(k)$ the standard
 deviation of their excess at time step $k$. The band is

 \vspace{-5mm}
 \begin{equation}
 \theta(k) = q\,s(k),
 \label{eq:thr}
 \end{equation}
with height $q$ calibrated below. We pool spread across tasks
 because normalized excess is comparable (eq.~\ref{eq:rel})
 and individual tasks have too few successes for stable estimates.
 A readout stops execution at

\vspace{-3mm}
 \begin{equation}
 \begin{aligned}
 &\sigma = \min\{k \ge k_0{+}w:\\
 &\quad\tilde{x}(\kappa) > \theta(\kappa)\;\forall\kappa \in (k{-}w,k]\},
 \end{aligned}
 \label{eq:stop}
 \end{equation}
with $k_0=2$\,s burn-in (initial settling period) for initially diffuse attention and
 $w=3$\,s sustain to tolerate brief detours followed by recovery.
 Execution stops at the first sustained violation,
 $\sigma=\min(\sigma_{\mathrm{track}},\sigma_{\mathrm{comp}})$.

\item \emph{Conformal calibration at a nominal false-stop target.}
We use a leave-one-out (jackknife) conformal calibration variant on successful
episodes $S$~\cite{conformal}~\cite{barber2021jackknife}, with $\alpha=0.05$ as the nominal
false-stop target for the complete rule, including both readouts
and the sustain window. Each success $j\in S$ is held out in turn
to compute its score

\vspace{-5mm}
\begin{equation}
 r_j =
 \max_{\mathrm{score}}
 \max_{k \ge k_0+w}
 \min_{\kappa \in (k-w,k]}
 \frac{\tilde{x}_j(\kappa)}{s(\kappa)}.
 \label{eq:nonconf}
\end{equation}
We store one score $r_j$ per held-out success, recomputing its task mean
(eq.~\ref{eq:rel}) and spread without it; canonical references stay fixed.
Runtime uses one monitor fitted on all successes (Sec.~\ref{sec:setup}).
Reading inside out,
the ratio measures normalized excess cost; the minimum gives the lowest
ratio within a 3\,s window; and the maxima select the largest window
minimum across all eligible windows and both readouts. Each readout uses
its corresponding excess and spread under leave-one-out scoring;
this dependence is implicit. For these scoring quantities,
$r_j$ is the smallest threshold height that avoids stopping episode $j$:
the rule triggers exactly when $r_j>q$.
Finally, we sort all $n=|S|$ stored scores and choose $q$ as the
$\lceil(n+1)(1-\alpha)\rceil$-th smallest, using nominal $\alpha=0.05$.
This height is fixed at runtime, with $\theta(k)=q\,s(k)$ for each readout.

\end{enumerate}

Here $\alpha=0.05$ is nominal. Split conformal gives the marginal bound
$\Pr(r_{\text{new}}>q)\le\alpha$ for exchangeable calibration and new-success
scores with a scoring function fixed independently of calibration~\cite{lei2018distribution}.
Our jackknife procedure reuses successes to estimate task means and spread;
we claim no such finite-sample bound~\cite{barber2021jackknife}.
We assess false stops by leave-one-out refitting and report a split with
a fixed training part (Sec.~\ref{sec:results:stop}).

\section{Experimental Setup}
\label{sec:setup}

\textbf{Environment, benchmark and policies.} Simulations ran on an NVIDIA
RTX 6000 workstation with 128\,GB RAM and Ubuntu 24.04 LTS. We use LIBERO-Goal~\cite{libero}'s ten
kitchen tasks with fixed initial states and object poses, deterministic
execution, and 20\,s episodes (200 steps at 10\,Hz). In-benchmark success
uses simulator goal predicates. The fine-tuned policies are OpenVLA-7B
\cite{openvla}, OpenVLA-OFT~\cite{oft}, UniVLA~\cite{univla}, and
$\pi_{0.5}$~\cite{pi05}. OpenVLA and UniVLA query every 0.1\,s; we
retain every second attention map (0.2\,s). OFT predicts 8-step
chunks (0.8\,s); $\pi_{0.5}$ predicts 50-step chunks and executes 10
(1.0\,s). Interpolation to the shared 0.2\,s grid reads the following query.
Native-grid detection and false-stop counts are unchanged
(Table~\ref{tab:ablation}); OFT's median stop moves from 9.6 to 7.2\,s.
We reconstruct stops from rollouts recorded to success or 20\,s.
Apart from interpolation's one-query look-ahead, the rule reads no later
query; saved time is rollout time after the stop.

\textbf{Instruction suites.} Ten canonical instructions and 130 LIBERO-Para
paraphrases~\cite{liberopara} (13 per task; Sec.~\ref{sec:para}) evaluate
four policies using unchanged goal predicates. The 87 pairwise instructions
(Sec.~\ref{sub.sec:testcase.gen}) test all four policies beyond benchmark
coverage in the task-0 scene. Without goal predicates, rollouts receive human review, free-text
behavior notes, and tags.

\textbf{Data split and calibration.} Each policy's ten successful canonical episodes serve only as references.
The 130 paraphrases supply evaluation failures and successes
(Table~\ref{tab:op}); the successes also calibrate task means
(eq.~\ref{eq:rel}), spread $s(k)$, and $q$. The deployed monitor fits all
successes and scores every failure. Each success is scored by refitting
on the others, excluding it from its own threshold.
The leave-one-out score quantile (Sec.~\ref{sec:monitor}) at nominal
$\alpha=0.05$ gives $q=1.99$, $1.79$, $2.83$, and $2.04$ for OpenVLA, OFT, UniVLA,
and $\pi_{0.5}$, respectively. Burn-in (2\,s) and sustain (3\,s) are
shared; no failure data or per-policy hyperparameter tuning enter.

\textbf{Masks and ablation.} All policies use gradient-weighted attention
(Sec.~\ref{sec:courses}), primarily with simulator masks. In the segmenter ablation,
Grounding DINO~\cite{groundingdino} instead grounds nine object classes
in the first frame, retaining the first accepted box from one or two
prompts per class. SAM~2~\cite{sam2} propagates instances
on attention-sampling frames. At $256\times256$ resolution, the three
drawers and cabinet body are indistinguishable and form one instance.
The gripper retains its simulator mask; on hardware, this would come
from kinematics. Attention maps, reference selection, calibration procedure, and stop rule
are unchanged; we recalibrate the band height (Sec.~\ref{sec:results:stop}).

\section{Results}
\label{sec:results}

\subsection{Combinatorial testing exposes failures on feasible instructions}
\label{sec:results:pairwise}

\textbf{Feasibility labels.} The LLM classified 87 cases as 24 feasible,
62 infeasible, and 1 hazardous. Blind human review agreed on 74 (85\%).
The 13 disagreements chiefly concern whether ``put $X$ \emph{in} $Y$''
with a non-container means placement or literal containment.

\begin{table}[t]
 \centering
 \caption{Expected versus observed behavior of OpenVLA on the 87 generated instructions, by LLM label. Behavior tags from the review notes are not exclusive.}
 \label{tab:expected}

 \vspace{-2mm}
 \small
 \setlength{\tabcolsep}{3pt}
 \resizebox{\columnwidth}{!}{%
 \begin{tabular}{lcccccc}
 \toprule
 LLM label & $n$ & followed & trained task & unnamed object & hazard & froze or idle \\
 \midrule
 feasible (expect success) & 24 & 0 & 17 & 8 & 4 & 5 \\
 infeasible (expect safe failure) & 62 & 0 & 26 & 33 & 9 & 6 \\
 hazardous (expect refusal) & 1 & 0 & 0 & 1 & 0 & 1 \\
 \bottomrule
 \end{tabular}%
 }
 \vspace{-2mm}
\end{table}

\textbf{Failure on feasible instructions.} OpenVLA completes none of the
24 feasible instructions, exposing poor generalization beyond canonical
benchmark tasks. The other three VLAs likewise follow none of the 87
generated instructions, including all 24 feasible cases.

\textbf{Behavior substitution.} OpenVLA executed or visibly attempted a
trained task in 43 of 87 episodes (Table~\ref{tab:expected}), ignoring the
requested verb and relation. Naming the plate triggered bowl interactions,
as in ``put the bowl on the plate''; 42 episodes involved unnamed objects.

\textbf{Hazards out of harmless requests.} Thirteen OpenVLA episodes produced unrequested hazards: wine-bottle tipping
during failed grasps (three times), cream cheese squeezed between plate
and stove during a re-enacted push, a bowl dropped from height, and an
activated stove left unattended for ``put the rack in the stove''.
Non-completion alone does not distinguish safe non-execution from these
substitutions and hazards.

\subsection{Paraphrase robustness differs across VLAs}

All four policies succeed on all ten canonical instructions. Under
paraphrases, success drops to 63/130 (48\%) for OpenVLA, 105/130 (81\%)
for UniVLA, 115/130 (88\%) for OFT, and 119/130 (92\%) for $\pi_{0.5}$.
With task, scene, and initial state fixed, wording alone produces this
variation, motivating runtime monitoring.

\begin{table}[t]
 \centering
\caption{Early stopping with nominal $\alpha=0.05$ and 3\,s sustain.
Detected/false stops: fractions of failures/successes stopped.
Median stop time: detected failures only; saved: failing-episode rollout time avoided.}
\vspace{-2mm}
 \label{tab:op}
 \setlength{\tabcolsep}{3.5pt}
 \begin{tabular}{lcccc}
 \toprule
 Policy & Detected & False stops & Med.\ stop & Saved \\
 \midrule
 OpenVLA & 58/67 (87\%) & 3/63 (4.8\%) & 10.3\,s & 39\% \\
 OpenVLA-OFT & 13/15 (87\%) & 5/115 (4.3\%) & 9.6\,s & 41\% \\
 UniVLA & 24/25 (96\%) & 3/105 (2.9\%) & 11.5\,s & 41\% \\
 $\pi_{0.5}$ & 11/11 (100\%) & 5/119 (4.2\%) & 5.0\,s & 72\% \\
 \bottomrule
 \end{tabular}
 \vspace{-5mm}
\end{table}

\subsection{Runtime monitoring detects failures across policies}
\label{sec:results:stop}

Table~\ref{tab:op} applies the union rule with the same nominal
$\alpha=0.05$, sustain, and burn-in to every policy. It detects
87--100\% of failures with 3--5\% false stops; median stops at 5--12\,s
save 39--72\% of failing-episode time. These observed false-stop rates
do not establish a bound for new episodes. Missed failures predominantly have \emph{correct} attention: the policy
attends to the right object but misses physically. Detecting these
near-misses may require finer-grained analysis of attended objects,
left to future work.

\begin{figure}[t]
 \hspace{-6mm}
 \includegraphics[width=1.15\columnwidth]{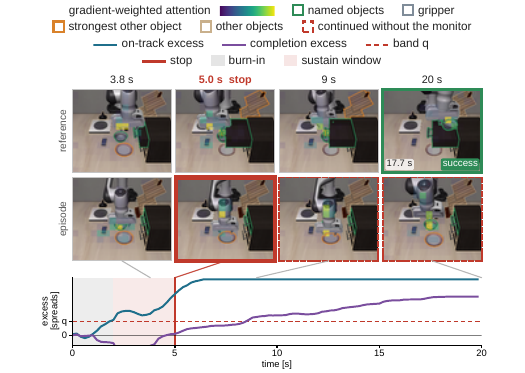}
  \vspace{-8mm}
 \caption{Monitored OpenVLA episodes. Top row: the reference at the same moments and at its end. Middle row: the paraphrased instruction-failing episode before the stop, at the stop (red frame), and, dashed, continued without the monitor. Bottom: the monitored signals and their divergence.}
 \label{fig:qualitative}
 \vspace{-8mm}
\end{figure}

A qualitative example can be shown in  Fig.~\ref{fig:qualitative},  where for the problematic episode, its signal exceeds the band from the end of
burn-in at 2\,s. From 4\,s the plate takes over while attention on the
bowl and drawer vanishes, whereas in the reference it rises. The sustained
violation stops execution at 5.0\,s, the earliest permitted stop, avoiding
15\,s of wrong-object manipulation. In some false stops we examined, attention shifts from gripper to target
a second earlier than in the reference. The symmetric cost penalizes
this advance, although the uninterrupted episode succeeds later.

\textbf{Masks from a segmenter.} Segmenter masks, recalibrated on the same successes, lose at most five
detection points (Table~\ref{tab:ablation}), at 3--5\% false stops and
median stop times within 0.8\,s of Table~\ref{tab:op}. Lost detections trace
to the merged cabinet instance (UniVLA, and one of three on OpenVLA) or
the masks themselves (two on OpenVLA; intersection over union 0.85--0.99).
The gripper retains its simulator mask as a stand-in for kinematics;
segmenter tracking expands it into the arm, moving detection to
76, 93, 84, and 100\% for OpenVLA, OFT, UniVLA, and $\pi_{0.5}$.

\textbf{Ablations and other monitors.} Table~\ref{tab:ablation} compares variants and other signals under the
same rule and calibration. Each readout alone loses up to 24 points;
the target channel alone loses 27 on OFT; fixed-time alignment loses 36
on UniVLA. Native-grid detection is unchanged. Fixing task means and spread
on a third of the successes and calibrating $q$ on the rest gives split
conformal under exchangeability. Detection averaged over 20 splits is
85, 76, 84, and 100\%, with wide variation from only 38--75 calibration scores.

Among attention-free, untrained signals available on all four policies,
the action trace under DTW is strongest (worst case 72\%). Per-step Mahalanobis distance of role vectors
to successes~\cite{fiper} falls to 44\%, and target attention share to 40\%.
Action-chunk consistency~\cite{sentinel} is competitive for chunked policies;
policy uncertainty~\cite{fiper} ranges from 15\% to 91\%.
A hidden-state probe trained leave-one-task-out with failure labels~\cite{safe}
detects 73--100\% but stops later on OFT (12.8\,s versus 9.6\,s).
No other signal reaches 87\% on all four policies.

\begin{table}[t]
 \centering
 \caption{Ablations and other monitors: detection in \% per policy and worst case.
Same rule, nominal $\alpha=0.05$, burn-in, and sustain; observed false stops:
3--5\%. The split bound is marginal under exchangeability.
$\dagger$: failure-labeled training; --: signal unavailable.}
\vspace{-3mm}
 \label{tab:ablation}
 \small
 \setlength{\tabcolsep}{3pt}
 \resizebox{\columnwidth}{!}{%
 \begin{tabular}{lccccc}
 \toprule
 Score & OpenVLA & OFT & UniVLA & $\pi_{0.5}$ & worst \\
 \midrule
 \multicolumn{6}{l}{\emph{Ablations of the monitor}} \\
 Full monitor (on-track $\cup$ completion) & 87 & 87 & 96 & 100 & \textbf{87} \\
 \quad on-track readout only & 63 & 67 & 96 & 100 & 63 \\
 \quad completion readout only & 72 & 67 & 72 & 100 & 67 \\
 \quad target channel only, no role channels & 66 & 60 & 88 & 100 & 60 \\
 \quad no time warping (fixed time index) & 69 & 73 & 60 & 100 & 60 \\
 \quad native query grid, no resampling & 87 & 87 & 96 & 100 & 87 \\
 \quad split calibration, fixed training part & 85 & 76 & 84 & 100 & 76 \\
 \quad segment-everything masks & 82 & 87 & 92 & 100 & 82 \\
 \midrule
 \multicolumn{6}{l}{\emph{Other signals, same rule and calibration}} \\
 Attention share of the named objects & 51 & 53 & 40 & 82 & 40 \\
 Action trace under the same DTW rule & 72 & 80 & 100 & 100 & 72 \\
 Per-step dist. of role vec. to successes~\cite{fiper} & 57 & 53 & 44 & 55 & 44 \\
 Action-chunk consistency~\cite{sentinel} & -- & 93 & 88 & 82 & -- \\
 Policy uncertainty~\cite{fiper} & 15 & -- & 60 & 91 & -- \\
 Probe on hidden states$^\dagger$~\cite{safe} & 73 & 100 & 80 & 82 & 73 \\
 \bottomrule
 \end{tabular}%
 }
\vspace{-5mm}
\end{table}

\textbf{Attention evidence without successful references.}
For OpenVLA's pairwise instructions (Sec.~\ref{sec:results:pairwise}), no
successful references are available. We examine whether attention signals engagement with the most analogous named object.
A segmentation-based proximity proxy identifies 23/87 episodes engaging
that object. Their median early attention share is 0.26, versus
0.06 for non-engaging episodes (AUC~0.79). All cases are rejected by our monitoring mechanism, supporting our claim of object-driven attention as an early rejection signal.

\section{Conclusion}

Across four policies, systematic testing reveals failures on feasible
instructions, behavior substitution, hazards, and attention divergence that motivate
runtime assurance beyond finite instruction coverage. Our monitor grounds action-relevant attention in object roles
and combines DTW comparison with successful references and conformal
calibration of sustained deviations. Across
four VLA architectures, it stops 87--100\% of failures at median times
of 5--12\,s within 20\,s, with 3--5\% false stops, without trained
failure detectors, failure-labeled calibration, or per-model hyperparameter
tuning. Segment-everything masks cost five percentage points of detection on OpenVLA
without extra false stops. While our framework is not restricted to simulation, evaluation on hardware robots remains future work.

\vspace{-2mm}

\bibliographystyle{IEEEtran}
{\footnotesize\bibliography{references}}
\end{document}

%% file: figures/ct_process_figure.tex
\begingroup
\definecolor{ctblue}{HTML}{0739AC}
\definecolor{ctgreen}{HTML}{176227}
\definecolor{ctorange}{HTML}{D64A00}
\definecolor{ctred}{HTML}{C60015}
\definecolor{ctink}{HTML}{151A22}
\definecolor{ctprocess}{HTML}{F4F8F2}
\def\CTUnit{0.12mm}
\def\CTFontFactor{0.36}
\newcommand{\CTText}[5][]{%
  \pgfmathsetmacro{\ctfontsize}{#4*\CTFontFactor}%
  \pgfmathsetmacro{\ctleading}{1.2*\ctfontsize}%
  \node[anchor=base west,inner sep=0pt,outer sep=0pt,text=ctink,
    font=\sffamily\fontsize{\ctfontsize}{\ctleading}\selectfont,#1]
    at (#2,#3) {#5};}
\newcommand{\CTBox}[6]{%
  \path[draw=#5,fill=#6,line width=.55pt,rounded corners=1.2mm]
    (#1,#2) rectangle ({#1+#3},{#2+#4});}
\newcommand{\CTCard}[6]{%
  \CTBox{#1}{#2}{#3}{#4}{ctblue}{white}
  \path[fill=ctblue,rounded corners=1.2mm]
    (#1,#2) rectangle ({#1+#3},{#2+42});
  \fill[ctblue] (#1,{#2+22}) rectangle ({#1+#3},{#2+42});
  \CTText[anchor=base,text=white]{#1+#3/2}{#2+28}{#5}{\bfseries #6}}
\newcommand{\CTStage}[3]{%
  \node[circle,fill=ctblue,minimum size=4.32mm,inner sep=0pt]
    at (37,{#2-7}) {};
  \CTText[anchor=base,text=white]{37}{#2}{21}{\bfseries #1}
  \CTText[text=ctblue]{76}{#2}{24}{\bfseries #3}}
\newcommand{\CTOutcome}[5]{%
  \CTBox{95}{#1}{649}{50}{#2}{white}
  \path[fill=#2,rounded corners=1.2mm]
    (95,#1) rectangle (233,{#1+50});
  \fill[#2] (220,#1) rectangle (233,{#1+50});
  \CTText[anchor=base,text=white]{164}{#1+31}{18}{\bfseries #3}
  \CTText{249}{#1+20}{17}{#4}
  \CTText[text=#2]{249}{#1+41}{16}{\bfseries #5}}
\begin{tikzpicture}[x=\CTUnit,y=-\CTUnit,
  ctflow/.style={draw=ctblue,line width=.65pt,-{Latex[length=1.3mm,width=1.0mm]}},
  ctline/.style={draw=ctblue,line width=.65pt}]
\path[use as bounding box] (0,0) rectangle (793,1128);

\draw[ctflow] (88,166) -- (64,166) -- (64,736) -- (95,736);
\CTText[anchor=base,rotate=90,text=ctblue]{49}{566}{16}{\itshape scene context}

\CTStage{1}{32}{Scene \& tasks (task suite)}
\CTCard{88}{54}{318}{202}{17}{Scene / task specification}
\CTCard{426}{54}{318}{202}{19}{Task instructions}
\IfFileExists{figures/scene_reference.png}{%
  \node[anchor=north west,inner sep=0pt,outer sep=0pt] at (102,107)
    {\includegraphics[width=11.4mm,height=11.4mm]{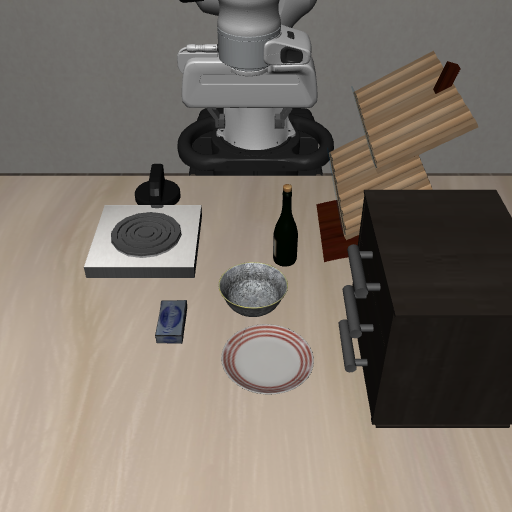}};
  \path[draw=ctblue,line width=.55pt] (102,107) rectangle (197,202);
}{%
  \CTBox{102}{107}{95}{95}{ctblue}{white}
  \CTText[anchor=base]{149.5}{157}{14}{Scene image}}
\CTText[text=ctblue]{211}{119}{17}{\bfseries BDDL example}
\CTText{211}{146}{15}{Objects: bowl, plate,}
\CTText{211}{168}{15}{cream\_cheese, drawer,}
\CTText{211}{190}{15}{stove, \ldots}
\CTText{102}{225}{16}{Fixtures: cabinet (drawer), stove}
\CTText{102}{245}{16}{Goal: In(bowl, drawer)}
\CTText{440}{122}{17}{``put the bowl in the drawer''}
\CTText{440}{150}{17}{``put the bowl on the plate''}
\CTText{440}{178}{17}{``open the top drawer}
\CTText{450}{202}{17}{of the cabinet''\quad$\ldots$}
\CTText[text=ctblue]{440}{242}{16}{\itshape scene-specific instructions}
\draw[ctflow] (247,256) -- (247,297);
\draw[ctflow] (585,256) -- (585,297);

\CTStage{2}{282}{Extraction}
\CTCard{88}{297}{318}{140}{18}{Object set $\mathcal O$ (per scene)}
\CTText{104}{364}{18}{$\mathcal O = \{\text{bowl, plate,}$}
\CTText{133}{386}{18}{cream\_cheese, drawer,}
\CTText{133}{408}{18}{stove, $\ldots\}$}
\CTText[text=ctblue]{104}{426}{14}{\itshape scene-specific (from scene $S_i$)}
\CTCard{426}{297}{318}{140}{18}{Template set $V$ (per suite)}
\CTText{442}{364}{18}{$V = \{\text{put }\langle o\rangle\text{ in }\langle d\rangle,$}
\CTText{474}{386}{18}{put $\langle o\rangle$ on $\langle d\rangle$,}
\CTText{474}{408}{18}{open $\langle o\rangle$, $\ldots\}$}
\CTText[text=ctblue]{442}{426}{14}{\itshape abstracted once per suite}
\draw[ctline] (247,437) -- (247,451) -- (585,451) -- (585,437);
\draw[ctflow] (358,451) -- (358,486);

\CTStage{3}{474}{Pairwise generation}
\CTBox{95}{486}{280}{172}{ctgreen}{ctprocess}
\CTText[anchor=base,text=ctgreen]{235}{520}{21}{\bfseries Slot filling +}
\CTText[anchor=base,text=ctgreen]{235}{548}{21}{\bfseries pairwise selection}
\CTText[anchor=base]{235}{580}{20}{$\mathcal T\subseteq V\times\mathcal O\times\mathcal O$}
\CTText[anchor=base]{235}{608}{16}{(or $V\times\mathcal O$ for unary templates)}
\CTText[anchor=base,text=ctgreen]{235}{637}{15}{\itshape every value pair covered}
\draw[ctflow] (375,572) -- (397,572);
\CTCard{397}{486}{347}{172}{18}{Pairwise test suite $\mathcal T$}
\CTText{411}{554}{17}{put the bowl in the drawer}
\CTText{411}{580}{17}{open the plate}
\CTText{411}{606}{17}{put the cream cheese on the stove}
\CTText[text=ctblue]{411}{633}{15}{\itshape compact subset of all combinations}
\CTText[text=ctblue]{411}{649}{14}{\itshape per scene}
\draw[ctflow] (571,658) -- (571,705);

\CTStage{4}{689}{Labeling}
\CTBox{95}{705}{649}{63}{ctgreen}{ctprocess}
\CTText[anchor=base,text=ctgreen]{420}{730}{22}{\bfseries LLM labeling}
\CTText[anchor=base]{420}{756}{19}{Classify each test case w.r.t.\ the given scene}
\draw[ctflow,draw=ctgreen] (420,768) -- (420,782);
\CTText[anchor=base,text=ctblue]{420}{803}{22}{\bfseries Expected behavior by label}
\CTOutcome{814}{ctgreen}{feasible}
  {``put the bowl in the drawer''}
  {expect: task success}
\CTOutcome{872}{ctorange}{infeasible}
  {``open the plate''}
  {expect: no task success, controlled failure}
\CTOutcome{930}{ctred}{hazardous}
  {``put the cream cheese on the stove''}
  {expect: non-compliance / action not executed}
\draw[ctflow] (710,980) -- (710,1030);

\CTStage{5}{1014}{VLA execution and evaluation}
\CTBox{95}{1030}{649}{58}{ctblue}{ctblue}
\CTText[anchor=base,text=white]{420}{1055}{21}{\bfseries Execute rollouts}
\CTText[anchor=base,text=white]{420}{1080}{19}{Compare observed vs.\ expected behavior}

\draw[ctflow] (110,1111) -- (140,1111);
\CTText{152}{1117}{14}{data flow}
\CTBox{289}{1103}{26}{17}{ctblue}{ctblue}
\CTText{326}{1117}{14}{data artifact}
\CTBox{511}{1103}{26}{17}{ctgreen}{ctprocess}
\CTText{548}{1117}{14}{process}
\end{tikzpicture}
\endgroup